\documentclass[10pt,twocolumn,letterpaper]{article}

\usepackage[pagenumbers]{cvpr} % To force page numbers, e.g. for an arXiv version

\usepackage{graphicx}
\usepackage{amsmath}
\usepackage{amssymb}
\usepackage{booktabs}
\usepackage{array}
\usepackage{tabularx}
\usepackage{pifont}% http://ctan.org/pkg/pifont
\usepackage{color, colortbl}

\usepackage{booktabs, multirow, makecell} % for borders and merged ranges
\usepackage{soul}% for underlines
\usepackage[table]{xcolor} % for cell colors

\usepackage{amssymb}% http://ctan.org/pkg/amssymb
\usepackage{pifont}% http://ctan.org/pkg/pifont

\definecolor{Gray}{gray}{0.9}

\usepackage[pagebackref,breaklinks,colorlinks]{hyperref}

\usepackage[capitalize]{cleveref}
\crefname{section}{Sec.}{Secs.}
\Crefname{section}{Section}{Sections}
\crefname{table}{Tab.}{Tabs.}
\Crefname{table}{Table}{Tables}

\def\cvprPaperID{xxxx} % *** Enter the CVPR Paper ID here
\def\confName{CVPR}
\def\confYear{2024}

\title{EgoSurgery-Tool: A Dataset of Surgical Tool and Hand Detection \\ from Egocentric Open Surgery Videos
}

\author{Ryo Fujii$^{1}$, Hideo Saito$^{1}$ and Hiroki Kajita$^{2}$\\
$^{1}$Keio University, $^{2}$Keio University School of Medicine \\ 
{\tt\small \{ryo.fujii0112, hs, jmrbx767\}@keio.jp}
}
\begin{document}

\makeatother
\twocolumn[{
\maketitle
\begin{center}
    \captionsetup{type=figure}
\includegraphics[width=\textwidth]{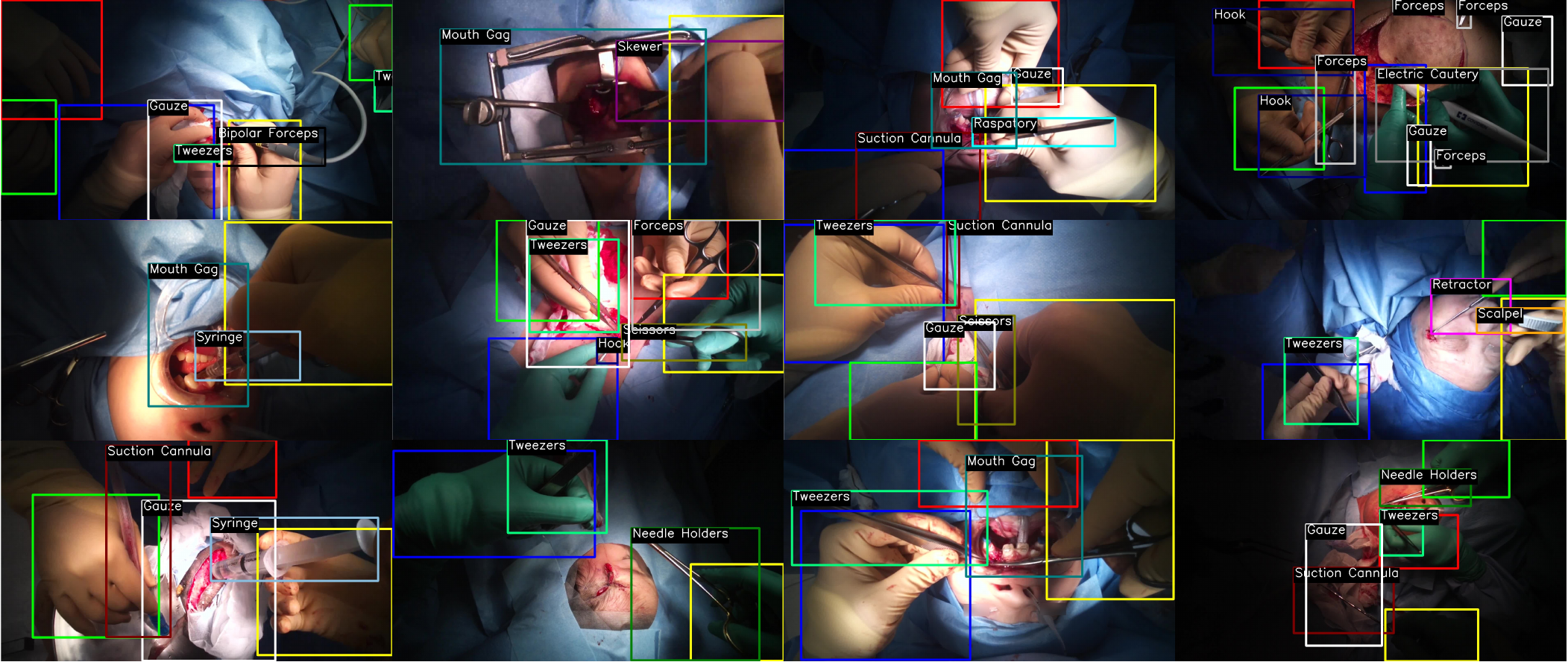}
    \captionof{figure}{Visualizations of EgoSurgery-Tool and ground truth annotations, for surgical tools,  \textcolor{blue}{own left hand}, \textcolor{yellow}{own right hand}, \textcolor{green}{other left hands}, and  \textcolor{red}{other right hands}.}
    \label{fig:egosurgery-examles}
\end{center}
}]
%===============================================================================

\begin{abstract}
Surgical tool detection is a fundamental task for understanding egocentric open surgery videos. However, detecting surgical tools presents significant challenges due to their highly imbalanced class distribution, similar shapes and similar textures, and heavy occlusion. The lack of a comprehensive large-scale dataset compounds these challenges. In this paper, we introduce EgoSurgery-Tool, an extension of the existing EgoSurgery-Phase dataset, which contains real open surgery videos captured using an egocentric camera attached to the surgeon’s head, along with phase annotations. EgoSurgery-Tool has been densely annotated with surgical tools and comprises over 49K surgical tool bounding boxes across 15 categories, constituting a large-scale surgical tool detection dataset. EgoSurgery-Tool also provides annotations for hand detection with over 46K hand-bounding boxes, capturing hand-object interactions that are crucial for understanding activities in egocentric open surgery. EgoSurgery-Tool is superior to existing datasets due to its larger scale, greater variety of surgical tools, more annotations, and denser scenes. We conduct a comprehensive analysis of EgoSurgery-Tool using nine popular object detectors to assess their effectiveness in both surgical tool and hand detection. The dataset will be released at \href{https://github.com/Fujiry0/EgoSurgery}{project page}.
\end{abstract}

\begin{table*}[tb]
\caption{Comparisons of EgoSurgery-Tool and existing datasets for surgical tool detection. \textit{OS} indicates open surgery.}
\begin{center}
\resizebox{\textwidth}{!}{
\begin{tabular}{lccccccc}
\toprule
Dataset  & Surgery type & Frames & Tool instances & Hand instances  & Tool categories & Hand categories & Tool Instances per frame \\
\cmidrule(l{2pt}r{3pt}){1-8} 
m2cai16-tool-locations~\cite{Jin2018WACV} & MIS & 2.8K  & 3.9K  &  & 7 & & 1.4   \\
Cholec80-locations~\cite{Shi2020Access} & MIS & 4.0K & 6.5K  &  & 7 & & 1.6 \\
AVOS dataset~\cite{Goodman2024JAMA}  & OS & 3.3K & 2.8K  & 6.2K  & 3  & 1  & 0.9 \\
\rowcolor{Gray} EgoSurgery-Tool (Ours) & OS & 15.4K & 49.7K & 46.3K & 15  & 4 & 3.2 \\
\bottomrule
\end{tabular}}
\end{center}
\label{tab:comparisons-with-existing-datasets}
\end{table*}

\section{Introduction}
Detecting surgical tools from an egocentric perspective in the operating room is fundamental task for the development of intelligent systems that can assist surgeons in real-time. For example, recognizing a tool can help prevent accidents, such as leaving gauze inside the body, by notifying surgeons. Recently, various approaches have been proposed for surgical tool detection, particularly in minimally invasive surgeries (MIS)\cite{Sarikaya2017TMI, Vardazaryan2018MICCAI, Jin2018WACV, Shi2020Access, Ali2022BBEngIV, Zia2023arXiv, FujiiICASS2024}. However, there have been few attempts to detect surgical tools in open surgery videos due to the limited availability of large-scale datasets. The existing surgical tool detection datasets for open surgery are either small\cite{Goodman2024JAMA} or not publicly available~\cite{FUJII2022AS}. In contrast, several datasets~\cite{Jin2018WACV, Shi2020Access, Raju2016report} have been released for MIS, driving advancements in learning-based algorithms. The absence of comparable large-scale datasets for open surgical tool detection has significantly impeded progress in achieving accurate tool detection within the open surgery domain. Challenges include dealing with surgical tools that exhibit a highly imbalanced, long-tailed distribution, have similar textures and shapes, and appear in occluded scenes, posing new challenges for many existing approaches.

Hand detection is an essential task for egocentric video analysis, where hand-object interaction (HOI) is crucial for action localization and understanding in activities of daily living. Several large-scale hand detection datasets have been proposed~\cite{ArpitBMVC2011, Bambach2015ICCV, Shan2020CVPR} for detecting hands in daily activities. Localizing hands is also vital for analyzing egocentric open surgery videos. However, there is little work on hand detection in the open surgery domain~\cite{Goodman2024JAMA, Zhang2021AMIA}, and only one small publicly available dataset exists~\cite{Goodman2024JAMA}. Training on existing hand datasets from daily activities does not transfer well to surgical hand detection due to significant differences in domain appearance, highlighting the need for a large-scale dataset.

With these motivations, we introduce EgoSurgery-Tool, a large-scale dataset captured from a camera attached to the surgeon's head, containing dense annotations for surgical tools and the surgeon's hand-bounding boxes. EgoSurgery-Tool is an extension of the recently introduced EgoSurgery-Phase~\cite{Fujii2024EgoSurgeryPhase}. We now elaborate on the unique characteristics and differences between the existing dataset~\cite{Goodman2024JAMA} and our proposed EgoSurgery-Tool dataset. Compared to the existing dataset~\cite{Goodman2024JAMA}, EgoSurgery-Tool offers several advantages: 1) it is the largest-scale dataset among tool and hand detection datasets in the open surgery domain in terms of the number of images and annotations; 2) it contains a greater variety of surgical tools; 3) it includes high-density scenes with numerous surgical tools; and 4) each hand annotation specifies hand identification (the camera wearer's left or right hand or another person's left or right hand). Our dataset is compared with existing related datasets in Table~\ref{tab:comparisons-with-existing-datasets}, and example images are shown in Figure~\ref{fig:egosurgery-examles}. Based on the proposed EgoSurgery-Tool dataset, we provide a systematic study on nine mainstream baselines.

\section{EgoSurgery-Tool Dataset}
The EgoSurgery-Phase dataset~\cite{Fujii2024EgoSurgeryPhase} consists of 21 videos covering 10 distinct surgical procedures, with a total duration of 15 hours, performed by 8 surgeons. EgoSurgery-Phase provides over 27K frames with phase annotations. However, EgoSurgery-Phase lacks sufficient information on surgical tools and hands. Therefore, we propose EgoSurgery-Tool, which includes additional annotations for surgical tools and hands on a subset of the existing EgoSurgery-Phase dataset. These annotations make EgoSurgery-Phase the only available dataset for multi-task learning of phase recognition, surgical tool detection, and hand detection. EgoSurgery-Phase is manually annotated by a group of annotators who were instructed for each task to ensure consistency across the dataset. The annotations were then inspected by expert surgeons to assess their quality. The rest of this section provides details on the annotations, benchmarking, and statistics of EgoSurgery-Tool.

\begin{table}[tb]
\caption{Comparison of datasets with respect to image distribution across various instance count ranges. We compute the number of images for each dataset within three count ranges.}
\begin{center}
\resizebox{\columnwidth}{!}{
\begin{tabular}{ccccc}
\toprule
 \multirow{2}{*}{Datasets}  &   \multirow{2}{*}{\begin{tabular}{c}\# Image  \\ (0-5 instances) \end{tabular} } &   \multirow{2}{*}{\begin{tabular}{c}\# Image  \\ (6-10 instances) \end{tabular} } &  \multirow{2}{*}{\begin{tabular}{c}\# Image  \\ (11-15 instances) \end{tabular} }\\
\\
\cmidrule(l{2pt}r{3pt}){1-4} 
m2cai16-tool-locations~\cite{Jin2018WACV} & 2,811 & 0  & 0 \\
\rowcolor{Gray} EgoSurgery-Tool & 6,128 & 8,803 & 506 \\
\bottomrule
\end{tabular}}
\end{center}
\label{tab:comparison-density}
\end{table}

\begin{figure}[tb]
 \centering
 \resizebox{\columnwidth}{!}{
\begin{minipage}{\hsize}
    \begin{center}
        \includegraphics[clip, width=\hsize]{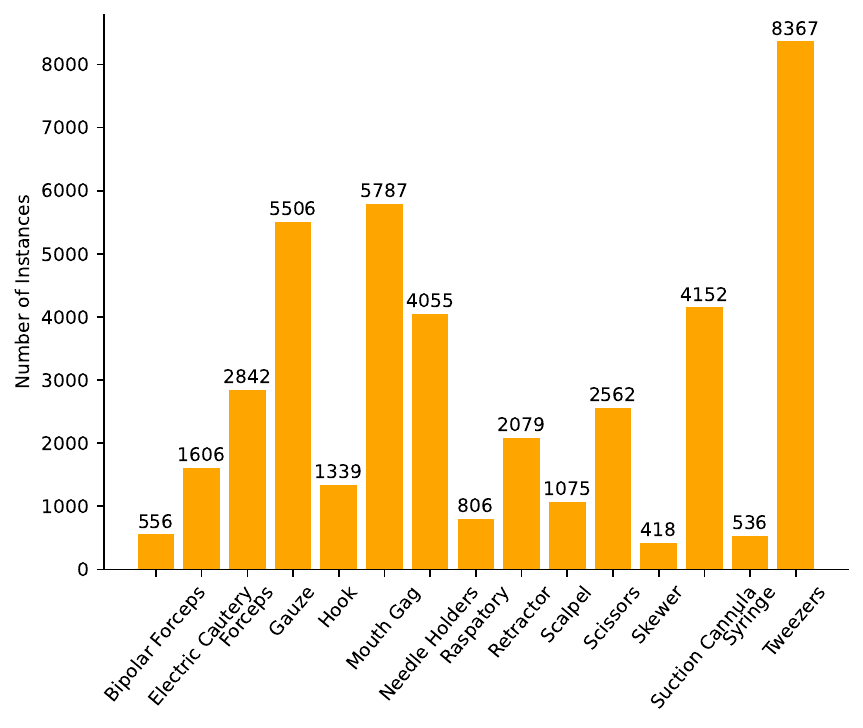}
    \end{center}
\end{minipage}
}
\caption{The distribution of surgical tool categories.}
 \label{fig:tools_dist}
\end{figure}

\begin{figure}[tb]
 \centering
 \resizebox{0.7\columnwidth}{!}{
\begin{minipage}{\hsize}
    \begin{center}
        \includegraphics[clip, width=\hsize]{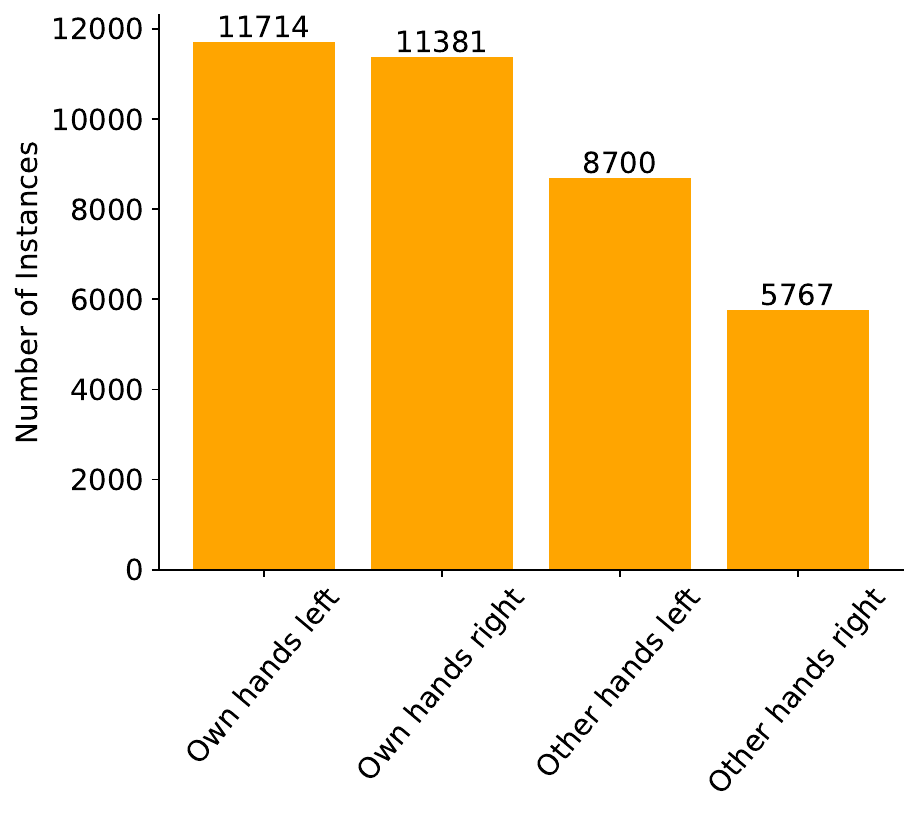}
    \end{center}
\end{minipage}
}
\caption{The distribution of hand categories.}
 \label{fig:hands_dist}
\end{figure}

\begin{table}[tb]
\caption{The number of instances per category in each set and the category distribution in the EgoSurgery-Tool dataset.}
\begin{center}
\begin{tabular}{c}
(a) The number of instances per surgical tool category.
\\
\resizebox{\columnwidth}{!}{
\begin{tabular}{cccccc}
\toprule
Class  & Train & Val & Test & Total & Dist.\\
\cmidrule(l{2pt}r{3pt}){1-6} 
Bipolar Forceps & 446 & 55 & 195 & 696 & 1.40\% \\
Electric Cautery  & 1,404 & 101 & 162 &  1,667 & 3.36\% \\
Forceps   &  2,534 & 154 & 3,375 & 6,063 & 1.22\% \\
Gauze & 4,596 & 455 & 1644 & 6,695 &  13.58\% \\
Hook & 1,045 & 147 & 157 & 1,349 & 2.72\%\\
Mouth Gag & 3,807 & 990 & 1,188 & 5,985 & 12.05\% \\
Needle Holders & 3,031 & 512 & 1,286 & 4,829 & 9.73\% \\
Raspatory & 654 & 76 & 84 & 814 & 1.64\% \\
Retractor & 2,079 & 0 & 325 & 2,404& 4.84\%  \\
Scalpel & 739 & 168 & 159 & 1,066 & 2.15\% \\
Scissors & 1,780 & 391 & 565 & 2,736 & 5.51\%\\
Skewer & 212 & 103 & 29  & 344 & 0.69\%\\
Suction Cannula & 3,134 & 509 & 768 &  4,411 & 8.88\% \\
Syringe & 344 & 96 & 141 & 581 & 1.17\% \\
Tweezers  & 6,467 & 950 & 2,595 & 10,012 & 20.16\% \\
\cmidrule(l){1-6}
Total & 32,272 & 4,707 & 12,673 & 49,652 & 100\% \\
\bottomrule
\end{tabular}}
\\
(b) The number of instances per hand category.
\\
\resizebox{\columnwidth}{!}{
\begin{tabular}{cccccc}
\toprule
Class  & Train & Val & Test & Total & Dist.\\
\cmidrule(l{2pt}r{3pt}){1-6} 
Own hands left & 8,704 & 1,505 & 3,834 & 14,043 & 30.3\% \\
Own hands right & 8,447 & 1,467 & 3,670 & 13,584 & 29.3\% \\
Other hands left & 6,542  & 1,079  & 3,412 & 11,033 & 29.3\% \\
Other hands right & 4,033 & 867 & 2,760 & 7,660 & 16.5\% \\
\cmidrule(l){1-6}
Total & 27,726 & 4,918 & 13,676 & 4,6320 & 100\% \\
\bottomrule
\end{tabular}}
\end{tabular}
\end{center}
\label{tab:instances-per-category}
\end{table}

\subsection{Data splits and statistic} 
\begin{figure}[tb] 
\centering
\includegraphics[width=\columnwidth]{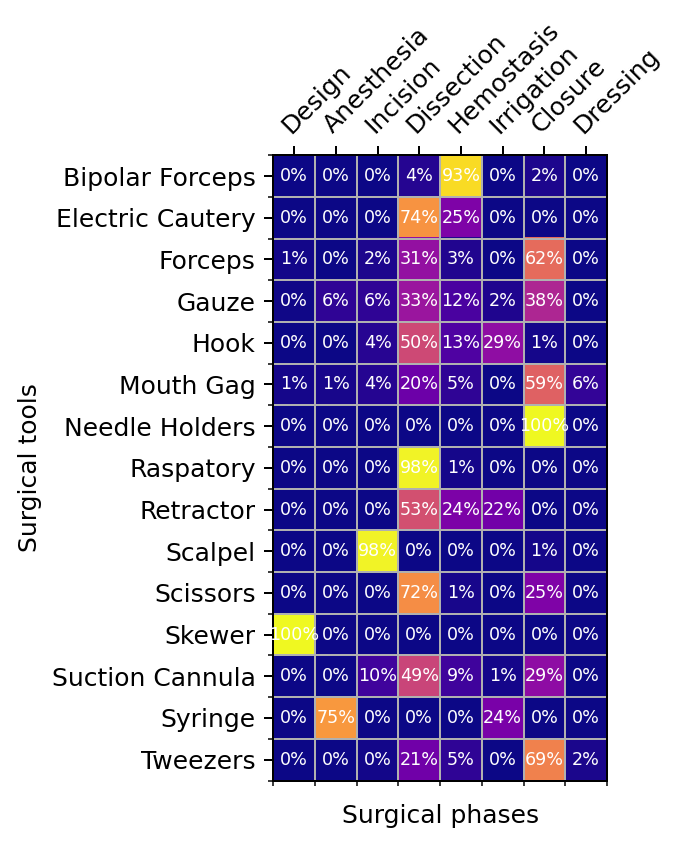}
\caption{Co-occurrence matrix between surgical tools and surgical phases.}
\label{fig:co_occurence_phase}
\end{figure}

We annotated 15 types of surgical tools and 4 types of hands in 15 videos from the EgoSurgery-Phase dataset. The proposed EgoSurgery-Tool dataset contains 15,437 high-quality images, annotated with 49,652 surgical tools and 46,320 hands. The distribution of surgical tools, shown in Figure~\ref{fig:tools_dist}, reveals a notable class imbalance. Figure~\ref{fig:hands_dist} shows The distribution of hand. Table~\ref{tab:comparison-density} shows the number of images within each instance count range (0-5, 6-10, 11-15). Our EgoSurgery-Phase dataset demonstrates higher density compared to the surgical tool detection dataset in MIS. The co-occurrence matrix between surgical tools and surgical phases is presented in Figure~\ref{fig:co_occurence_phase}. Along the Y-axis are the given surgical tools, and the X-axis enumerates conditional phases. Each element represents the conditional probability that a phase occurs when a surgical tool is used. For example, when a scalpel appears in a frame, that frame belongs to the incision phase with a probability of 0.98. Surgical tool information might be helpful for surgical phase recognition. EgoSurgery-Tool is divided into training, validation, and test sets at the video level, ensuring that all frames of a video sequence appear in one specific split. The 15 video sequences are split into 10 training, 2 validation, and 3 test videos for consistency with the standard evaluation of other relevant datasets, resulting in 9,657 training, 1,515 validation, and 4,265 test images. The number of instances per category in each set is shown in Table~\ref{tab:instances-per-category}.

\section{Experiments}
\subsection{Experimental setups} 
We compare nine popular object detectors: Faster R-CNN
(2015)~\cite{Ren2015NeurIPS}, RetinaNet (2017)~\cite{Lin2017ICCV}, Cascade R-CNN (2018)~\cite{Cai2018CVPR}, CenterNet (2019)~\cite{Zhou2019arXiv}, Sparse R-CNN (2021)~\cite{Sun2021CVPR}, VarifocalNet (2021)~\cite{Zhang2021CVPR}, Deformable-DETER (2021)~\cite{Zhu2021ICLR}, DDQ (2023)~\cite{Zhang2023CVPR}, and DINO (2023)~\cite{Zhang2023ICLR}.   We use the MMDetection~\cite{mmdetection} for the implementation. We fine-tune models with pre-trained on MS-COCO~\cite{Lin2014ECCV}. For
a fair comparison, we select the algorithm’s backbones to
have a similar number of parameters. We use the COCO evaluation procedure and report $AP$, $AP_{50}$, and $AP_{75}$~\cite{Lin2014ECCV}. Because each detector is calibrated differently, setting a comparable detection confidence threshold is impractical. Therefore, we evaluate all the detectors by using confidence $10^{-8}$.

\begin{figure*}[tb] 
\centering
\includegraphics[width=\linewidth]{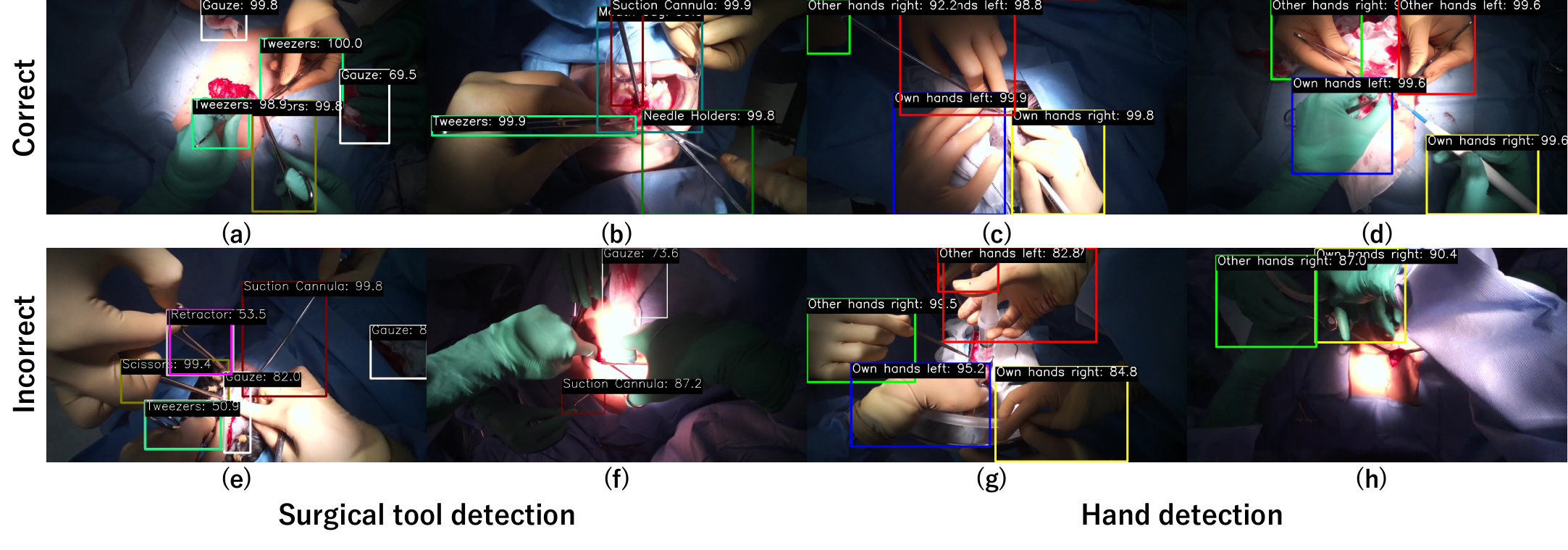}
\caption{Qualitative results for the object detection challenge. The first column shows correct detections, while the second column shows incorrect cases.}
\label{fig:qualitative-detection-results}
\end{figure*}

\begin{table}[tb]
\caption{Performance of object detection methods on the EgoSurgery-Tool. The best performance is shown in bold.}
\begin{center}
\begin{tabular}{c}
(a) Surgical tool detection performance.
\\
\resizebox{\columnwidth}{!}{
\begin{tabular}{cccc}
\toprule
Methods   & $AP$ & $AP_{50}$ & $AP_{75}$ \\
\cmidrule(l{2pt}r{3pt}){1-4} 
Faster R-CNN~\cite{Ren2015NeurIPS} & 37.7 & 55.8 & 43.3  \\
RetinaNet~\cite{Lin2017ICCV}  &  36.2 & 53.0 & 39.8\\
Cascade R-CNN~\cite{Cai2018CVPR} & 38.8 & 55.7 & 44.6 \\
CenterNet~\cite{Zhou2019arXiv}  & 42.4 & 60.2 & 46.8  \\
Sparse R-CNN~\cite{Sun2021CVPR}& 37.0 & 55.1 & 41.8\\
VarifocalNet~\cite{Zhang2021CVPR} & \textbf{45.8} & \textbf{63.3} & \textbf{51.1}  \\
Deformable-DETR~\cite{Zhu2021ICLR} & 30.0 & 46.3 & 34.0 \\
DDQ~\cite{Zhang2023CVPR}  & 43.2 & 59.1 & 48.7\\
DINO~\cite{Zhang2023ICLR}  &  39.7  & 56.7 & 43.5\\
\bottomrule
\end{tabular}
}
\\
(b) Hand detection performance.
\\
\resizebox{\columnwidth}{!}{
\begin{tabular}{cccc}
\toprule
Methods   & $AP$ & $AP_{50}$ & $AP_{75}$ \\
\cmidrule(l{2pt}r{3pt}){1-4} 
Faster R-CNN~\cite{Ren2015NeurIPS} & 55.3 & 80.4 & 62.3   \\
RetinaNet~\cite{Lin2017ICCV}   & 57.1 & 81.9 & 62.9  \\
Cascade R-CNN~\cite{Cai2018CVPR} & 55.5 & 80.7 & 61.4\\
CenterNet~\cite{Zhou2019arXiv}  & 56.6 & 78.5 & 63.3  \\
Sparse R-CNN~\cite{Sun2021CVPR} & 55.4 & 78.7 & 60.9 \\
VarifocalNet~\cite{Zhang2021CVPR} & \textbf{59.4} & \textbf{82.1} & 65.3   \\
Deformable-DETR~\cite{Zhu2021ICLR} & 54.1 & 78.6 & 59.2  \\
DDQ~\cite{Zhang2023CVPR}  & 58.3 & 73.5  & 60.8 \\
DINO~\cite{Zhang2023ICLR}  & 58.8  & 80.2 & \textbf{65.6} \\
\bottomrule
\end{tabular}}
\end{tabular}
\end{center}
\label{tab:comparison-existing-object-detection-methods}
\end{table}

\begin{figure}[tb] 
\centering
\includegraphics[width=\columnwidth]{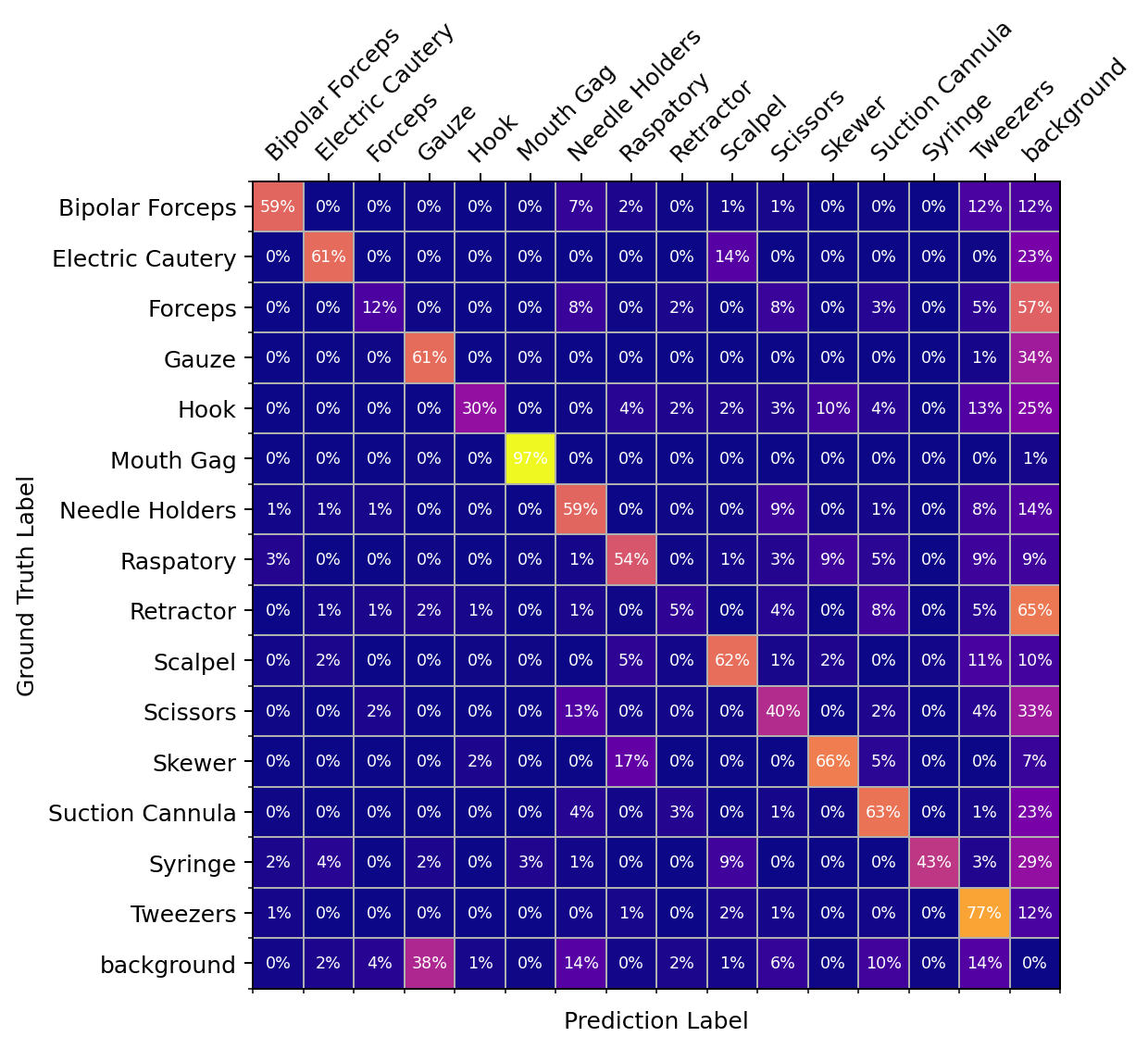}
\caption{Confusion matrix of surgical tool detection model.}
\label{fig:confusion_matrix_tools}
\end{figure}

\begin{table}[tb]
\caption{Left: Faster-RCNN hand detection performance comparison between the existing hand detection dataset, EgoHands, and our dataset. Right: Pretrained Faster-RCNN hand detection performance with fine-tuning on our dataset, separated by training order.}
\vspace{-1em}
\begin{center}
\resizebox{\columnwidth}{!}{
\begin{tabular}{cc}
\begin{tabular}{cc}
\toprule
  Training data & $AP$ \\
  \cmidrule(l{2pt}r{3pt}){1-2} 
EgoHands & 8.9   \\
\rowcolor{Gray} Ours & \textbf{55.3} \\
\bottomrule
\end{tabular}
&
\begin{tabular}{cc}
\toprule
  Pre-training dataset & $AP$ \\
  \cmidrule(l{2pt}r{3pt}){1-2} 
ImageNet & 50.7 \\
COCO &  \textbf{55.3} \\
COCO, EgoHands & 52.1 \\
\bottomrule
\end{tabular}
\end{tabular}}
\end{center}
\label{tab:comparison-with-difference-datasets}
\end{table}

\subsection{Quantitative results}
We present the results of nine mainstream object detection algorithms in Table~~\ref{tab:comparison-existing-object-detection-methods}. For surgical tool detection, among all methods, the recent VarifocalNet achieves the highest performance in terms of the $AP$ metric for surgical tool detection tasks. VarifocalNet also consistently outperforms other detectors in terms of $AP_{75}$ and $AP_{50}$, indicating its superior ability to estimate the correct bounding box sizes. The superiority of VarifocalNet is attributed to its dense object detection capability, enabling it to detect objects at small scales and under heavy occlusion. For hand detection, VarifocalNet outperforms other object detection methods in terms of $AP$ and $AP_{75}$. In terms of $AP_{50}$, DINO achieves the best performance.

The confusion matrix for the standard object detection method, Faster R-CNN, is shown in Figure~~\ref{fig:confusion_matrix_tools}. We observe that tools with similar textures and shapes are often misclassified (\eg, scissors and needle holders). Additionally, tools with many varieties of appearances are confused with backgrounds (\eg, forceps, gauze, and retractors).

We compare the hand detection performance of different training data and pre-training data settings using Faster R-CNN in Table\ref{tab:comparison-with-difference-datasets}. Training with our EgoSurgery-Tool dataset significantly outperforms training with the existing hand dataset, EgoHands, which was collected in a daily living setting. Despite the vast quantity of annotated data in EgoHands, models trained solely on EgoHand perform substantially worse compared to those trained with our EgoSurgery-Tool, suggesting a significant domain transfer problem related to the characteristics and representation of hands in a surgical environment. We also explored the performance of hand detection with different pre-training data. Pre-training with COCO achieves the best performance. Due to the significant domain gap, pre-training with the existing hand detection dataset, EgoHands, degrades performance.

\subsection{Qualitative results}
Figure~\ref{fig:qualitative-detection-results} presents qualitative results for Faster-RCNN using IoU thresholds of $0.5$. The model successfully detects surgical tools in (a, b) and hands wearing different colors of surgeons' gloves in (c, d) across a variety of surgery types. Examples of detection failures are shown in (e)-(h). Heavy occlusion (e, h), poor lighting conditions (f), and similar shapes and textures between categories (e, g) cause these incorrect detections.

\section{Conclusion}
To address the lack of a large-scale dataset in the open surgery domain, we introduce EgoSurgery-Tool, an egocentric open surgery video dataset captured from a camera attached to the surgeon's head, including bounding box annotations for surgical tools and hands. We conducted extensive evaluations of recent object detection methods on this new benchmark dataset. We believe the dense annotations of EgoSurgery-Tool will foster future research in video understanding within the open surgery domain.

%===============================================================================
\section*{ACKNOWLEDGMENT}
This work was supported by JSPS KAKENHI Grant Number 22H03617.

%===============================================================================

% \clearpage

{\small
\bibliographystyle{ieee_fullname}
\bibliography{ref}
}

\end{document}